\documentclass[runningheads]{llncs}
\usepackage[T1]{fontenc}
\usepackage{multirow}
\usepackage{booktabs}
\usepackage{graphicx,verbatim}
\usepackage{amsmath}
\usepackage{amssymb}
\usepackage[table]{xcolor}
\usepackage{fontawesome5}
\definecolor{bestgreen}{RGB}{230,242,230}

\begin{document}

\title{B-MIM: Biased Masked Image Modeling for Generalizable Segmentation of Fine-Grained Anatomical Structures}

\titlerunning{B-MIM for Fine-Grained CT Segmentation}

\author{Sebastián González\inst{1}\and
Karen Sanchez\inst{2}\and
José M. Saavedra\inst{3}\and
Marcelo Pizarro\inst{3}\and
Bernard Ghanem\inst{2}
}

\authorrunning{S. González et al.}
\institute{Universidad de Chile, Santiago, Chile \and
King Abdullah University of Science and Technology (KAUST), Saudi Arabia \and
Universidad de los Andes, Santiago, Chile\\
\email{sebagonzalez@ug.uchile.cl}}

  
\maketitle
\begin{abstract}
Self-supervised pretraining enables transferable representations for medical imaging, yet most CT encoders remain biased toward coarse semantic understanding, limiting their sensitivity to fine-grained anatomical structures such as vessels or small tumors. In this paper, we introduce Biased Masked Image Modeling (B-MIM), a modification of the iBOT objective that stochastically reduces global semantic alignment to prioritize local patch reconstruction. This bias encourages the encoder to capture high-frequency morphological details and structural continuity.
We curate a multi-institutional CT abdominal dataset of 9,955 filtered studies from 17 public sources and pretrain a 3D Swin Transformer backbone using B-MIM. 
Across inter-dataset experiments on liver vessel segmentation, the proposed encoder improves topological fidelity (clDice) and achieves competitive Dice scores in tumor segmentation, compared to fully fine-tuned baselines, despite updating only a fraction of the parameters. Our results suggest that reducing global semantic pressure during pretraining enhances generalization to intricate anatomical structures.


\end{abstract}

\section{Introduction}

Computerized Tomography (CT) scans are a critical, non-invasive diagnostic tool in modern medicine that combine multiple X-ray images to produce detailed, 3D views of internal structures. They are essential for rapidly diagnosing, staging, and monitoring diseases—particularly cancer, cardiovascular issues, and trauma—guiding treatment, and reducing the need for exploratory surgery \cite{Buzug2011}. In addition, machine learning is increasingly the preferred method for extracting and interpreting information from these kinds of images, as evidenced by a vast body of publications on machine learning for medical imaging \cite{Varoquaux2022}. 

The de facto standard for a machine learning model follows an encoder-decoder architecture. The encoder aims to extract relevant information from images, text, or any input modality, while the decoder uses those representations to accomplish a target task. It is now a common practice to use encoders trained under a self-supervised regimen because of their capability to learn from unlabeled data \cite{ericsson:2022,dinov3,pe}. 

Learning transferable representations facilitates adaptation to diverse tasks, allowing the model to drastically reduce its number of learnable parameters, which is highly important in low-resource environments. That is why self-supervised learning strategies have become critical for making semantic encoders available. In this vein, models like MoCo \cite{he:2020:moco}, SimCLR \cite{chen:2020} and BYOL \cite{grill:2020} have contributed to the state of the art on image representation,  which is dominated by the DINO encoder family  \cite{dinov2,dinov3}. 
 
The medical imaging domain has benefited from advances in computer vision and machine learning. Thus, we are seeing encoders for X-ray images, such as RadDINO \cite{raddino}, RayDINO \cite{raydino} and Google's CXR \cite{healthfoundation}. However, for the CT domain, there are still a few proposals.  Here, VoCo \cite{voco-1,voco-2}, a self-supervised strategy for 3D medical imaging using contrastive learning, showed improvements across diverse downstream tasks using CT or MRI.

However, existing pretrained models and encoders for medical imaging (e.g., Total Segmentator \cite{wasserthal:2023}, Segment-Any-Muscle \cite{onno:2025}, UKBOB \cite{ukbob}) and encoders like VoCo \cite{voco-1,voco-2} focus mainly on coarse-grained tasks, such as macro-organ segmentation, with comparatively less emphasis on fine-grained anatomical structures, such as vessels or small tumors.  Fine-grained structures play a critical role in disease prevention, early detection, disease monitoring, and treatment planning. For instance,  liver vessel segmentation \cite{crlm_simpson:2024} is critical to plan liver resection, and the early detection of small tumors is crucial to cancer treatment. 

Therefore, this work proposes a CT encoder designed to enhance representations of fine-grained anatomical structures, such as vessels and small tumors. We evaluate our encoder on liver vessel and small-tumor segmentation tasks, showing its potential for inter-dataset transfer. For instance, we observed up to a $18\%$ gain in vessel segmentation when the proposal is applied to an out-of-distribution dataset. Thus, our contributions are summarized as follows:
\begin{itemize}
    \item We curate a multi-institutional dataset from 17 public sources into a standardized cohort of 9,955 CT studies (1,993,194 2D slices) with consistent LPS orientation and anatomical cropping for 3D self-supervised learning.
    \item We introduce the Biased Masked Image Modeling (B-MIM) objective, which prioritizes local patch reconstruction over global semantic alignment. By leveraging the multi-scale nature of the 3D Swin Transformer, this strategy encourages the encoder to capture high-frequency morphological details, specifically addressing the topological continuity required for fine-grained structures.
    \item We demonstrate that our pre-trained backbone serves as a fixed-feature extractor, achieving competitive results on specialized tasks (e.g., vessel and tumor segmentation) with only 16k trainable parameters.   
\end{itemize}

\section{Related Works}
\label{sec:related}
\textbf{Self-Supervised Learning (SSL):} Modern artificial intelligence models are based on an \textit{encoder-decoder} scheme, where the encoder is responsible for properly representing an image, text, or audio signal into a semantic feature space (a.k.a. latent space). 
In recent years, we have seen the emergence of increasingly powerful image encoders \cite{chen:2020,grill:2020,dinov2,dinov3,pe}. For an encoder to be truly effective, it is essential to make use of as much available data as possible, even though this data often lacks proper labeling. To this end, self-supervision plays a key role, enabling the model to learn from diverse images and understand discriminative structures that facilitate downstream inference. Furthermore, the absence of labels during the encoder training stage can help models generalize better avoiding underfitting to specific classes. 


In this vein, DinoV3 \cite{dinov3} is one of the most effective visual encoders; it incorporates iBOT \cite{zhouimage}, a mechanism to extract information from small local structures. The idea is that a patch encoding can be reconstructed with contextual information when the image patch is occluded. Thus, we leverage iBOT loss to allow the model to focus on small structures.  

In the context of medical imaging, we still find a few works proposing medical visual encoders. Most available encoders are designed for the X-ray domain. For instance, META released RayDino \cite{raydino}, extending the DINO architecture using a collection of chest X-rays. RayDino is a large visual encoder trained by self-supervision on 873,000 chest X-rays. RayDINO shows state-of-the-art performance across many radiology tasks, from classification and dense segmentation to text generation, and provides an in-depth analysis of population, age, and sex biases of the model. The authors suggest that SSL allows patient-centric AI to be useful in clinical workflows and to interpret X-rays holistically. In the same direction, RadDino \cite{raddino} was proposed by Microsoft, and CXR by Google \cite{healthfoundation}. 


VoCo \cite{voco-1,voco-2} is a pretrained methodology for 3D medical images that proposes the Volume Contrast (VoCo) framework to leverage contextual position priors during pre-training. This method groups crops from different regions while enforcing feature discrepancy among them. As the X-Ray encoders, VoCo is not focused on fine-grained tasks, which is our main goal throughout this work.

Pretrained encoders favor generalization and make adaptation easier. These properties are critical in medical imaging. However, there is a significant gap to cover, especially in tasks involving small anatomical structures using image modalities other than X-ray. Therefore, this work proposes a CT encoder specifically designed to improve sensitivity to fine-grained anatomical structures, such as vessels and small tumors. 
    
\section{Methodology}
\label{sec:methodology}
Our methodology comprises three stages: i) the preparation and curation of the training dataset, ii) the design of the encoder architecture, and iii) an evaluation methodology based on the inter-dataset generalization property. Following, we describe each of the proposed components.

\subsection{Data Curation and Large-Scale Pre-processing} 
To train an encoder for cross-dataset segmentation of fine-grained abdominal structures, we curated a large-scale multi-institutional CT dataset by aggregating 17 publicly available sources in both DICOM and NIfTI formats. The initial collection underwent a rigorous automated filtering and normalization pipeline, including standardized orientation and automated abdominal cropping to remove non-relevant anatomical regions and ensure consistent anatomical representation.

\textbf{Filtering Criteria.} We applied a standardized selection protocol to both data formats. For DICOM series, we excluded localizer scans and instances with null orientation headers or inconsistent slice orientations (e.g., mismatched first- and middle-slice headers). Furthermore, we restricted the cohort to adult patients (age $\ge 18$ or unspecified) and enforced a minimum volumetric depth of 20 slices per series after abdominal cropping. A critical requirement for inclusion was the simultaneous presence of the liver and at least one kidney, to ensure that the model learns consistent abdominal spatial relationships.

\textbf{Dataset Composition.} From 13,784 abdominal CTs originally compiled, the final filtered cohort comprised 9,955 studies (cases), totaling 1,993,194 2D slices. 
The following summary highlights the primary data streams:
\textit{DICOM Cohort:} A total of 5,613 studies were selected from sources such as rsna2023-abdominal (3,147 studies), TCGA-KIRC (362), and CT-Colonography (831), totaling 1,529,954 2D slices.
\textit{NIfTI Cohort:} 4,342 studies, totaling 463,240 slices, were retained from datasets including AMOS (2,338 cases), Ab\-do\-menCT-1K (1,061), and KiTS23 (480). Notably, Ab\-do\-menCT-1K itself integrates subsets from LiTS, MSD Spleen, and MSD Pancreas.

\textbf{Normalization and Cropping.} All volumes were normalized to the LPS (Left-Posterior-Superior) orientation. To eliminate non-relevant anatomical noise, we performed automated abdominal cropping using TotalSegmentator \cite{wasserthal:2023}  to identify landmarks. The vertical bounds were defined from the superior aspect of the liver to the inferior pole of the L4 vertebra or the kidneys, whichever was lower. TotalSegmentator is used exclusively for automated Region-of-Interest localization during preprocessing. No downstream task annotations (e.g., vessels or tumors) are used at this stage, ensuring that the cropping procedure does not introduce supervision related to the evaluation tasks.

The curated large-scale abdominal CT cohort will be publicly released upon acceptance. Due to space constraints, the complete list of source datasets and corresponding references is available in the repository: \\ \url{https://github.com/ksanchez84/B-MIM}

\subsection{Biased Masked Image Modeling (B-MIM)}

\begin{figure}
    \centering   \includegraphics[width=1\linewidth]{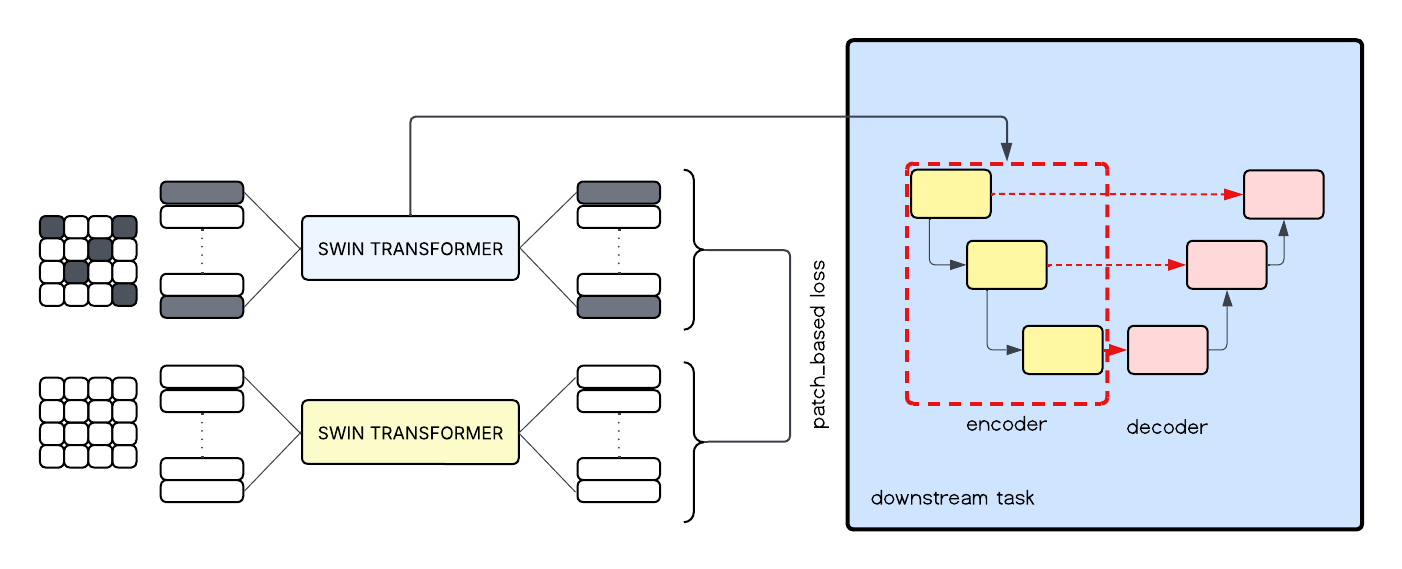}
    \caption{General scheme of the proposed CT encoder and how it is integrated into a downstream segmentation task.}
    \label{fig:encoder}
\end{figure}

Figure \ref{fig:encoder} shows the general scheme of our proposed encoder and how it is integrated into a downstream model. Following, we describe each of the B-MIM components.

\subsubsection{Backbone Architecture.} We employ a 3D Swin Transformer \cite{liu2021swin,swinunetr-1,swinunetr-2} as the backbone for our encoder. The input 3D CT volumes are partitioned into non-overlapping patches, and a hierarchical architecture computes self-attention within shifting windows. This setup allows the model to capture both local textures and long-range spatial dependencies, providing the sensitivity required to represent detailed, fine-grained structures with high fidelity.

\subsubsection{Pre-training Objective.} The encoders are pre-trained using the iBOT framework \cite{zhouimage}, framed as a cross-view self-distillation task. Given two augmented views of a scan, the student network learns to match the teacher network's output using a dual distillation loss. Following iBOT, the teacher network has the same architecture as the student and is updated as an exponential moving average of the student parameters. The global \(L_{\mathrm{CLS}}\) and masked-token \(L_{\mathrm{MIM}}\) distillation losses are defined as
\begin{equation}
L_{\mathrm{CLS}} = H(P_t^{\mathrm{CLS}},P_s^{\mathrm{CLS}}), \qquad
L_{\mathrm{MIM}} = \frac{1}{|\mathcal{M}|}\sum_{i\in\mathcal{M}} H(P_{t,i},P_{s,i}),
\end{equation}

where \(H\) denotes cross-entropy and \(\mathcal{M}\) the set of masked patches.
However, to enhance the segmentation of fine-grained structures, we propose a Biased-MIM objective, where the total loss $L_{total}$ is formulated as:

\begin{equation}
L_{total} = \mathbb{I}_{p} \cdot L_{CLS} + L_{MIM},
\qquad \mathbb{I}_{p}\sim\mathrm{Bernoulli}(p).
\end{equation}

where $L_{CLS}$ is the self-distillation loss between [CLS] tokens, $L_{MIM}$ is the distillation between masked patch tokens, and $\mathbb{I}_{p} \in \{0, 1\}$ is a Bernoulli random variable with probability $p$, sampled independently at each training iteration (per batch), determining whether the global distillation loss $L_{CLS}$ is applied for that batch.
In contrast to deterministic loss weighting, stochastic omission alters the optimization dynamics by intermittently removing global supervision altogether, thereby reducing co-adaptation between global and local objectives and promoting robustness to fine-scale morphological variation.

\subsubsection{Rationale for the Bias.} The standard iBOT formulation treats both losses with equal priority ($p=1$). However, to enhance the segmentation of fine-grained structures, we set $p < 1$ to intentionally reduce the frequency of global feature alignment. By stochastically omitting global supervision, we encourage the student network $f_s$ to rely more heavily on reconstructing masked tokens via $L_{MIM}$. This bias encourages the encoder to learn high-resolution local geometries and continuity, which are essential for thin structures, rather than over-relying on coarse global semantic features.

\subsubsection{Parameter-Efficient Downstream Adaptation}

To evaluate the representational quality of the pre-trained embeddings, we employ a parameter-efficient adaptation strategy. Instead of conventional full fine-tuning, which is computationally expensive and prone to forgetting anatomical priors in small datasets, we attach a lightweight nnU-Net decoder to the frozen Swin-3D encoder.

\subsubsection{Encoder-Decoder Integration.}
For adaptation, we tested the following three primary configurations:
\begin{enumerate}
    \item \textit{Feature Concatenation}: Merging multi-scale embeddings from the pretrained encoder into the corresponding decoder levels. The encoder is kept largely frozen, with only 16,344 parameters updated during adaptation. We call this setting \textbf{B-MIM-A}.
    \item \textit{Full Finetuning}: We performed full finetuning on the same architecture in order to assess differences from updating all encoder parameters. We call this setting \textbf{B-MIM-B}.
    \item \textit{LoRA Adaptation (Optional/Comparative)}: We integrated Low-Rank Adaptation \cite{lora2022} to further refine the bottleneck without unfreezing the backbone. We call this setting \textbf{B-MIM-C}.
\end{enumerate}


\subsection{Generalization-based Evaluation}
We evaluate our proposed encoder on two fine-grained downstream tasks: liver vessel segmentation and small-tumor segmentation using abdominal CTs as input. In both cases, we use the CRLM dataset (197 studies) \cite{crlm_simpson:2024} for training and validation.
Given that one of our main goals is to evaluate generalization, we use two datasets for testing, not used during pretraining: IRCAD (20 studies) \cite{ircad_soler20103d} and MSD (303 studies) \cite{msd_simpson:2019}.
The encoder is pretrained in a fully self-supervised manner on the curated abdominal CT cohort, without access to vessel or tumor annotations. For downstream evaluation, models are trained exclusively on CRLM (5-fold cross-validation), and evaluated on IRCAD and MSD as external test sets. This protocol ensures patient-level separation and prevents label leakage between pretraining and downstream evaluation.
        

\section{Results and Discussion}
\label{sec:results}

The 3D Swin Transformer backbone was pretrained using the B-MIM objective for 200 epochs with a batch size of 6. We adopted the standard iBOT optimization configuration with AdamW, using a linear warm-up for the first 10 epochs followed by a cosine learning rate schedule decaying from $7.5 \times 10^{-4}$ to $2 \times 10^{-6}$.

For downstream segmentation, models were trained using 5-fold cross validation (k = 5), with a batch size of 2 for 200 epochs, following the default nnU-Net training protocol. Optimization was performed using SGD with a polynomial learning rate scheduler (initial learning rate = $1 \times 10^{-2}$, exponent = 0.9) and a constant weight decay of $3 \times 10^{-5}$. We report mean Dice and clDice scores across folds. The encoder remained frozen unless otherwise specified. All experiments were conducted on an NVIDIA RTX 6000 Ada Generation (48 GB) GPU.




\subsection{Segmentation Performance and Inter-Dataset Generalization}

A key property of the proposed approach is its ability to generalize across diverse clinical domains without unfreezing the backbone.
All models were trained on CRLM, and we report mean Dice and clDice across folds. Generalization was evaluated on IRCAD and MSD, which exhibit distinct acquisition protocols and anatomical variability.
In all experiments presented in this section, B-MIM results use $p = 0.3$, which provided a stable trade-off between global semantic alignment and local morphological reconstruction in preliminary validation experiments.

\textbf{Vessel Segmentation}. As shown in Table 1, B-MIM consistently enhances topological fidelity (clDice) in vessel segmentation across datasets. Notably, under domain shift from CRLM to IRCAD, B-MIM-C improves clDice from 0.491 (Swin-nnUNet) to 0.582, corresponding to a gain of +0.091 absolute points (+18.5\% relative improvement). This improvement in topological fidelity is substantially larger than the corresponding Dice variation, highlighting the benefit of the biased objective for preserving vascular continuity across datasets.

Further, even the lightweight B-MIM-A configuration, with only 16,344 trainable encoder parameters, achieves competitive Dice and clDice scores on both IRCAD and MSD, demonstrating that fine-grained morphological representations can generalize across datasets with minimal adaptation.

\begin{figure}
    \centering   \includegraphics[width=1\linewidth]{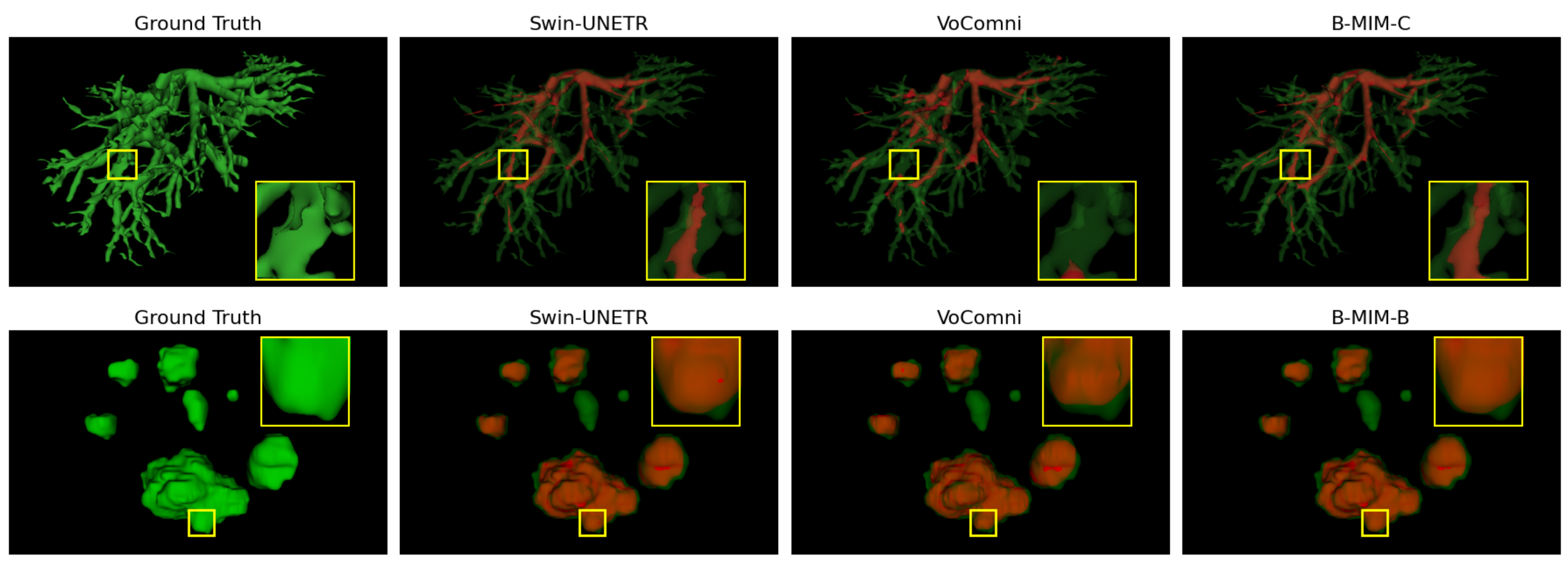}
    \caption{Qualitative comparison for fine-grained liver vessel and tumor segmentation. The top row shows liver vessel predictions, where B-MIM better preserves the continuity of thin vascular branches and reduces local fragmentation compared with Swin-UNETR and VoComni. The bottom row shows tumor segmentation, where B-MIM provides predictions that more closely match the ground truth in the highlighted regions. Zoomed areas emphasize local differences in fine-grained structure delineation.}
    \label{fig:vessel_result}
\end{figure}

Figure~\ref{fig:vessel_result} provides qualitative examples for both liver vessel and tumor segmentation. For vessel segmentation, B-MIM-C better preserves the continuity of thin vascular branches and exhibits less fragmentation than Swin-UNETR and VoComni. For tumor segmentation, B-MIM-B produces delineations that more closely follow the ground-truth boundaries in the highlighted regions.

\textbf{Parameter Efficiency.} A core contribution of this work is to achieve high-fidelity segmentation with minimal architectural updates. A key observation from Table 1 is the trade-off between trainable parameters and generalization performance. While B-MIM-B updates 5,716,410 parameters and achieves strong tumor performance, B-MIM-A maintains competitive vessel segmentation with only 16,344 trainable parameters. This suggests that the pretrained encoder captures transferable fine-grained morphological priors that require only lightweight adaptation.

\textbf{Tumor Segmentation.} In tumor segmentation, performance differences are more variable. B-MIM-B achieves the best Dice in this setting, likely due to the higher variability in tumor morphology and size. Unlike vessels, which exhibit consistent tubular structure, tumor morphology exhibits higher inter-patient variability in size, shape, and contrast patterns, making representation transfer inherently more challenging.

\begin{table*}[t]
\centering
\footnotesize
\caption{Segmentation performance across datasets for different encoder configurations and adaptation strategies. Dice and clDice are reported for liver vessels, while Dice is reported for tumors.}
\label{tab:segmentation_results}

\setlength{\tabcolsep}{3.1pt}
\renewcommand{\arraystretch}{1.05}

\begin{tabular}{lccccccccc}
\toprule
\multirow{2}{*}{Model} 
& \multicolumn{6}{c}{Liver Vessels} 
& \multicolumn{3}{c}{Tumors} \\
\cmidrule(lr){2-7}
\cmidrule(lr){8-10}
& \multicolumn{2}{c}{CRLM}
& \multicolumn{2}{c}{IRCAD}
& \multicolumn{2}{c}{MSD}
& CRLM & IRCAD & MSD \\
\cmidrule(lr){2-3}
\cmidrule(lr){4-5}
\cmidrule(lr){6-7}
\cmidrule(lr){8-8}
\cmidrule(lr){9-9}
\cmidrule(lr){10-10}
& Dice & clDice
& Dice & clDice
& Dice & clDice
& Dice & Dice & Dice \\
\midrule

Swin-nnUNet
& \cellcolor{bestgreen}\textbf{0.746}
& \cellcolor{bestgreen}\textbf{0.780}
& 0.400
& 0.491
& 0.598
& 0.624
& 0.646
& 0.572
& 0.374 \\

VoCo\_SSL
& 0.616
& 0.645
& 0.382
& 0.437
& 0.488
& 0.480
& 0.650
& 0.369
& 0.288 \\

VoComni
& 0.632
& 0.662
& 0.455
& 0.494
& 0.572
& 0.588
& \cellcolor{bestgreen}\textbf{0.666}
& 0.500
& \cellcolor{bestgreen}\textbf{0.423} \\

\midrule

B-MIM-A
& 0.744
& 0.777
& 0.413
& 0.500
& \cellcolor{bestgreen}\textbf{0.611}
& \cellcolor{bestgreen}\textbf{0.631}
& 0.631
& 0.497
& 0.357 \\

B-MIM-B
& 0.745
& \cellcolor{bestgreen}\textbf{0.780}
& 0.401
& 0.490
& 0.604
& 0.630
& 0.651
& \cellcolor{bestgreen}\textbf{0.578}
& 0.395 \\

B-MIM-C
& \cellcolor{bestgreen}\textbf{0.746}
& 0.779
& \cellcolor{bestgreen}\textbf{0.492}
& \cellcolor{bestgreen}\textbf{0.582}
& 0.567
& 0.595
& 0.637
& 0.506
& 0.269 \\
\bottomrule
\end{tabular}
\end{table*}


\section{Conclusions}
\label{sec:conclusions}

We introduced B-MIM, a biased masked pretraining objective designed to enhance representations of fine-grained anatomical structures in CT. By reducing global semantic pressure during self-supervised training, the proposed encoder improves topological fidelity in vessel segmentation while remaining parameter-efficient during downstream adaptation. Our experiments show that vessel structures, which exhibit a consistent tubular morphology, benefit noticeably from this representation bias and achieve improved cross-dataset transfer in our evaluation. In contrast, tumor segmentation remains more challenging due to higher variability in size, shape, and appearance. Future work will involve a more comprehensive evaluation across fine-grained tasks stratified by lesion size and complexity, a systematic analysis of the sensitivity to \(p\), and extension to additional modalities such as MRI.

\bibliographystyle{splncs04}
\bibliography{references}

\end{document}